\documentclass[letterpaper,10pt,conference]{ieeeconf}
\IEEEoverridecommandlockouts
\usepackage{graphicx}
\usepackage{amsmath,amssymb}
\usepackage{booktabs,array,tabularx}
\usepackage{cite,url}
\usepackage[hidelinks]{hyperref}
\hypersetup{pdftitle={ForceBench: Benchmarking Force Integration for Robotic Manipulation},pdfauthor={Xuan Han, Zinuo Jin, Tianlin Li, Yunfan Lou, Xin Liu, Yuchen Xue, Yuanpeng He, Boxian Yao, Shuhan Zhan, Ruihai Wu, Zhibo Pang}}
\usepackage{xcolor}
\usepackage{tikz}
\DeclareRobustCommand{\figcircled}[1]{\tikz[baseline=(n.base)]{\node[draw,circle,line width=0.35pt,minimum size=8pt,inner sep=0pt,outer sep=0pt,font=\normalfont\fontsize{6}{6}\selectfont,text height=4.2pt,text depth=0pt] (n) {#1};}}
\usepackage{microtype}
\newcolumntype{L}[1]{>{\raggedright\arraybackslash}p{#1}}
\newcolumntype{Y}{>{\raggedright\arraybackslash}X}
\AddToHook{env/thebibliography/begin}{\emergencystretch=3em\hyphenpenalty=100}
\graphicspath{{figures/}}
\makeatletter
\providecommand{\bstctlcite}[1]{\@bsphack\@for\@citeb:=#1\do{\edef\@citeb{\expandafter\@firstofone\@citeb}\if@filesw\immediate\write\@auxout{\string\citation{\@citeb}}\fi}\@esphack}
\makeatother
\newcommand{\bench}{ForceBench}
\newcommand{\pifive}{\ensuremath{\pi_{0.5}}}

\title{\LARGE\bf
ForceBench: Benchmarking Force Integration\\for Robotic Manipulation
}
\author{%
\authorblockN{Xuan Han$^{1*}$, Zinuo Jin$^{1*}$, Tianlin Li$^{1*}$, Yunfan Lou$^{3,1*}$, Xin Liu$^{1}$,\\
Yuchen Xue$^{1}$, Yuanpeng He$^{1}$, Boxian Yao$^{1}$, Shuhan Zhan$^{1}$,
Ruihai Wu$^{2\dagger}$, Zhibo Pang$^{1,4\dagger}$}
\authorblockA{$^{1}$PKU-PI Lab, Peking University\quad
$^{2}$University of California, Berkeley\\
$^{3}$National University of Singapore\quad
$^{4}$ePyBot Intelligence}%
\thanks{$^{*}$Equal contribution. $^{\dagger}$Corresponding authors.}%
}

\makeatletter
\IEEEaftertitletext{%
\begin{minipage}{\textwidth}
\centering
\def\@captype{figure}
\includegraphics[width=0.83\textwidth,trim=208bp 41bp 192bp 38bp,clip]{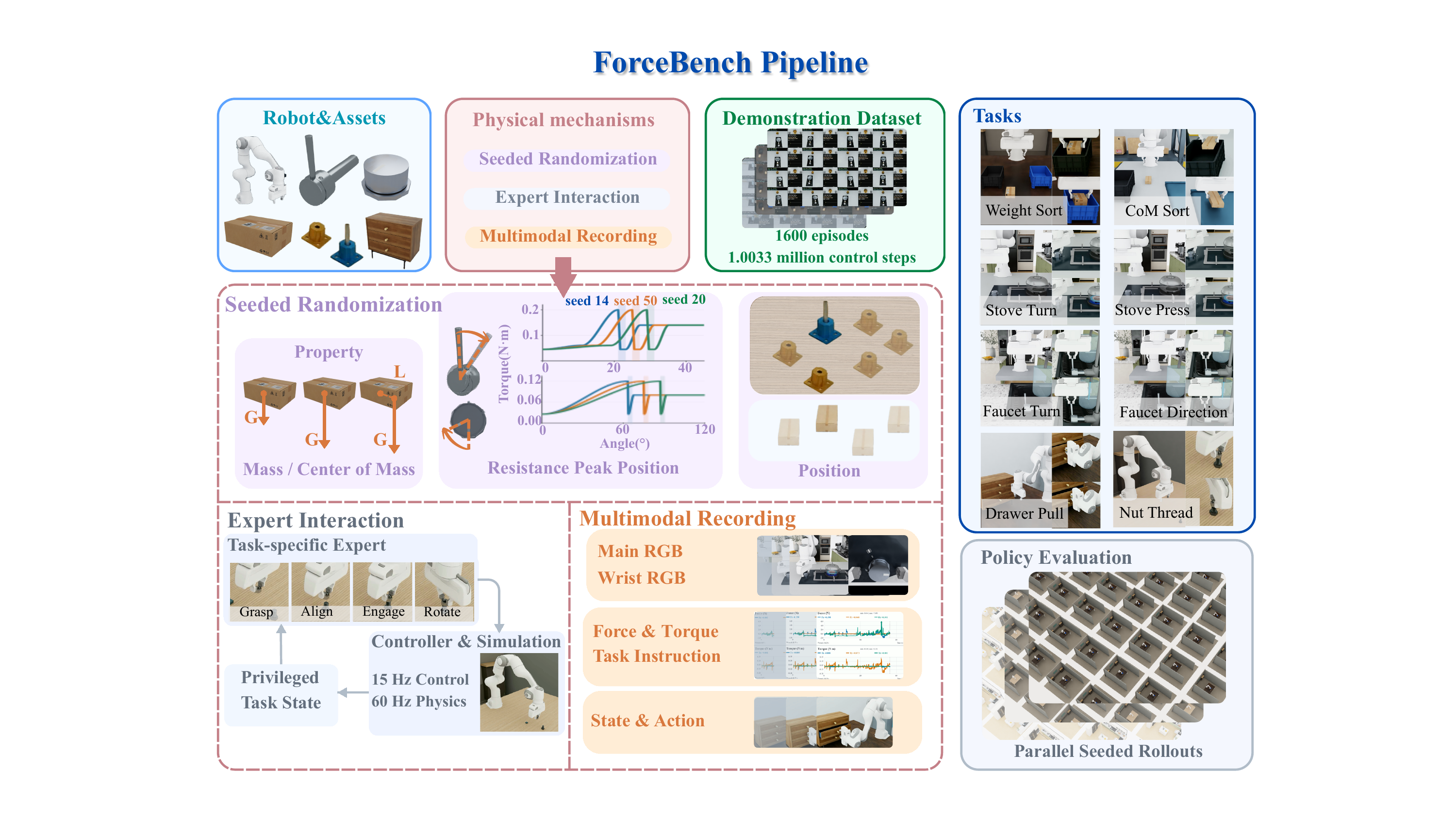}
\caption{Overview of the ForceBench pipeline. Robot and task assets are combined with parameterized physical mechanisms to construct eight manipulation tasks. Seeded instantiation varies object mass, center of mass, resistance profiles, and initial poses to generate reproducible task instances. Task-specific scripted experts use privileged simulator state to execute staged interactions through a shared controller, with control at 15 Hz and physics simulation at 60 Hz. External and wrist RGB images, six-axis force/torque measurements, task instructions, robot states, and actions are recorded synchronously, yielding 1,600 successful demonstrations. Policies are evaluated through parallel rollouts initialized with shared seeds, enabling consistent comparisons of force integration designs across physical-property inference, resistance-dependent interaction, and contact-rich assembly.}
\label{fig:pipeline}
\end{minipage}
\vspace{8pt}%
}
\makeatother

\begin{document}
\bstctlcite{IEEEexample:BSTcontrol}
\maketitle
\thispagestyle{empty}
\pagestyle{empty}
\raggedbottom

\begin{abstract}
Force feedback provides vision-language-action (VLA) policies with
physical observations that can guide action selection and execution.
However, differences in tasks, demonstrations, and control settings
across existing systems hinder reliable comparisons of force integration
designs. We introduce \bench{}, a simulation benchmark for evaluating
six-axis wrist-force integration under a unified protocol.
\bench{} combines eight tasks spanning physical-property inference,
resistance-dependent interaction, and contact-rich assembly with
1,600 synchronized demonstrations and task-specific success criteria.
We compare one no-force baseline and six force integration designs on
a shared \pifive{} backbone. Prompt Fusion achieves the highest observed mean simulation success
of 35.31\%, compared with 23.25\% for the baseline, and improves
success rates across all three task categories.
Predictive and temporal gains in simulation concentrate on
physical-property inference, while both temporal designs
underperform the baseline on assembly.
Real-world evaluation of eight systems across five tasks supports
the main qualitative trends: Prompt Fusion and Dual-Path Fusion
achieve 63.00\% mean success, compared with 40.00\% for the baseline,
but only Prompt Fusion improves every task.
These findings support \bench{}'s practical value for identifying
the benefits and limitations of force integration across
manipulation demands.
The project page is available at
\url{https://forcebench-platform.github.io/}.
\end{abstract}

\begingroup\small\noindent
\par\endgroup
\medskip
\section{Introduction}
\label{sec:intro}

Vision-language-action (VLA) models connect visual and language pretraining
to robotic action, enabling a shared backbone to learn diverse manipulation
skills~\cite{kim2024openvla,black2024pi0,intelligence2025pi05visionlanguageactionmodelopenworld,kim2025oft}.
Physical interaction additionally requires responding to properties and
conditions revealed through action, such as object mass and resistance.
Learning-based policies increasingly exploit such feedback, extending
diffusion-based visuomotor policies~\cite{chi2023diffusion} with compliance-
or force-aware learning~\cite{hou2024acp,liu2025factr,xue2025rdp} and
incorporating force into VLA models. Wrist force provides a six-dimensional
observation of these interactions, combining three force and three torque
components. These measurements can refine a visually specified motion or
inform decisions about conditions that appearance alone does not reveal,
supporting both action selection and execution.

Existing systems incorporate force into action generation through adaptive
fusion of force tokens with vision--language
features~\cite{yu2025forcevlaenhancingvlamodels} or temporal force encoding
with high-frequency feedback~\cite{li2026favla,wang2026phaforce}.
Force also supports reactive execution through contact prediction that
regulates feedback usage~\cite{he2024foar}, while joint action--wrench
modeling supports reinforcement-learning post-training~\cite{deng2026facet}.
Other work removes the force sensor
at deployment by distilling force knowledge into vision~\cite{zhao2026fdvla}.
Although standardized data and benchmarks have matured for vision-based
manipulation~\cite{fang2026domino,lan2025autobio},
differences in tasks, demonstrations, and control settings across
force-aware systems hinder direct comparisons. Moreover, aggregate
success rates can obscure differences across physical-property inference,
resistance-dependent interaction, and contact-rich assembly. We therefore
ask how force integration designs can be reliably compared under a unified
protocol, and how their benefits vary across manipulation demands.

To address this gap, we introduce \bench{}, a simulation benchmark
dedicated to comparing six-axis wrist-force integration in robotic
manipulation (Fig.~\ref{fig:pipeline}). \bench{} combines parameterized
physical conditions, 1,600 synchronized multimodal demonstrations,
a shared observation--action interface, and evaluation on task instances
initialized with shared seeds. Seven tasks vary physical properties and
resistance conditions to examine behavior beyond object appearance;
a complementary contact-rich assembly task tests geometrically
constrained execution. Task-specific success criteria and category-level
metrics support comparison across these manipulation demands. We further
assess the practical relevance of the benchmark through real-world
experiments.

Using this framework, we compare a no-force baseline and six force
integration designs on a shared \pifive{} backbone, using the same task
demonstrations and training protocol. The six designs comprise three
current-sample conditioning interfaces---Token Fusion, Prompt Fusion,
and Cross-Attention---a Token Fusion variant with auxiliary future-force
prediction, and two temporal reference systems, Temporal Prompt and
Dual-Path Fusion. ForceVLA provides an external comparison.
Prompt Fusion achieves the highest observed mean simulation success
of 35.31\%, compared with 23.25\% for the no-force baseline, and provides
the most consistent improvements across task categories.
In simulation, predictive supervision and temporal representations
provide task-dependent gains that concentrate on physical-property
inference, while both temporal designs underperform the baseline on
contact-rich assembly.

Real-world evaluation of eight systems across five tasks supports the
main qualitative trends. Prompt Fusion raises mean success from 40.00\%
to 63.00\%, tying Dual-Path Fusion overall while exceeding the no-force
baseline on every task. Both temporal designs achieve higher
physical-property inference success than current-sample Prompt Fusion,
but lower assembly success than the no-force baseline. These results
show that similar overall success rates can conceal different tradeoffs
across manipulation demands, supporting \bench{}'s practical value for
comparing force integration designs.

Our contributions are threefold:
\begin{itemize}
    \item \textbf{A unified benchmark for force integration.}
    \bench{} provides eight tasks with parameterized physical conditions,
    1,600 synchronized multimodal demonstrations, and shared evaluation
    with task- and category-level outcomes for six-axis wrist-force
    integration.

    \item \textbf{A systematic comparative study.}
    We evaluate observation conditioning, auxiliary predictive
    supervision, and temporal force representations on a shared VLA
    backbone, characterizing their benefits and limitations across
    physical-property inference, resistance-dependent interaction,
    and contact-rich assembly.

    \item \textbf{Real-world evaluation of benchmark findings.}
    We evaluate eight systems across five real-world tasks in 800 trials.
    The results support the main qualitative trends observed in
    simulation and reveal task-dependent tradeoffs that overall success
    rates alone can obscure.
\end{itemize}

\section{Related Work}
\label{sec:related}

\textbf{Force-Aware Policies for Robotic Manipulation.}
Token-based approaches incorporate force or tactile observations into
policy representations~\cite{yu2025forcevlaenhancingvlamodels,huang2025tactilevla}.
Text-based conditioning includes proprioceptive discretization in
\pifive{}~\cite{intelligence2025pi05visionlanguageactionmodelopenworld}
and stage-specific force prompts in ForceVLA2~\cite{li2026forcevla2}.
ManipForce and FAVLA use cross-attention~\cite{lee2025manipforce,li2026favla},
while FoAR regulates visual--force fusion using predicted
contact~\cite{he2024foar}.
TA-VLA and ForceFlow incorporate future torque or force
prediction~\cite{zhang2025tavla,forceflow2026}, and MoSS combines
sensory streams with cross-modal attention and predictive
supervision~\cite{lee2026moss}.
FAWAM couples action--force prediction with online residual
correction~\cite{he2026fawam}; our auxiliary prediction head is used
only during training.
ForceMimic links force-aware demonstration collection to hybrid
force--position imitation learning~\cite{liu2024forcemimic}.
Recent work extends tactile policies with predictive world
models~\cite{lou2026dreamtac},
while ContactWorld examines their visuo-tactile
representations~\cite{zhang2026contactworld}.

\textbf{Benchmarking and Evaluation of Robotic Manipulation.}
RLBench, Meta-World, CALVIN, and LIBERO provide task
suites and evaluation protocols for vision-based
manipulation~\cite{james2019rlbenchrobotlearningbenchmark,yu2021metaworldbenchmarkevaluationmultitask,mees2022calvin,liu2023libero}.
RoboCasa, RoboTwin 2.0, DOMINO, and AutoBio extend coverage to
household, bimanual, dynamic, and laboratory
settings~\cite{nasiriany2024robocasalargescalesimulationeveryday,chen2025robotwin20scalabledata,fang2026domino,lan2025autobio}.
SIMPLER studies correspondence between simulated and real-world
policy evaluations~\cite{li2024evaluatingrealworldrobotmanipulation},
and UniVTAC benchmarks visuo-tactile perception and
interaction~\cite{chen2026univtac}.
ForceBench complements these resources with shared demonstrations
and a unified protocol for comparing six-axis wrist-force integration
across physical-property inference, resistance-dependent interaction,
and contact-rich assembly.

\begin{figure}[!t]
\centering
\includegraphics[width=\linewidth,trim=226bp 34bp 252bp 52bp,clip]{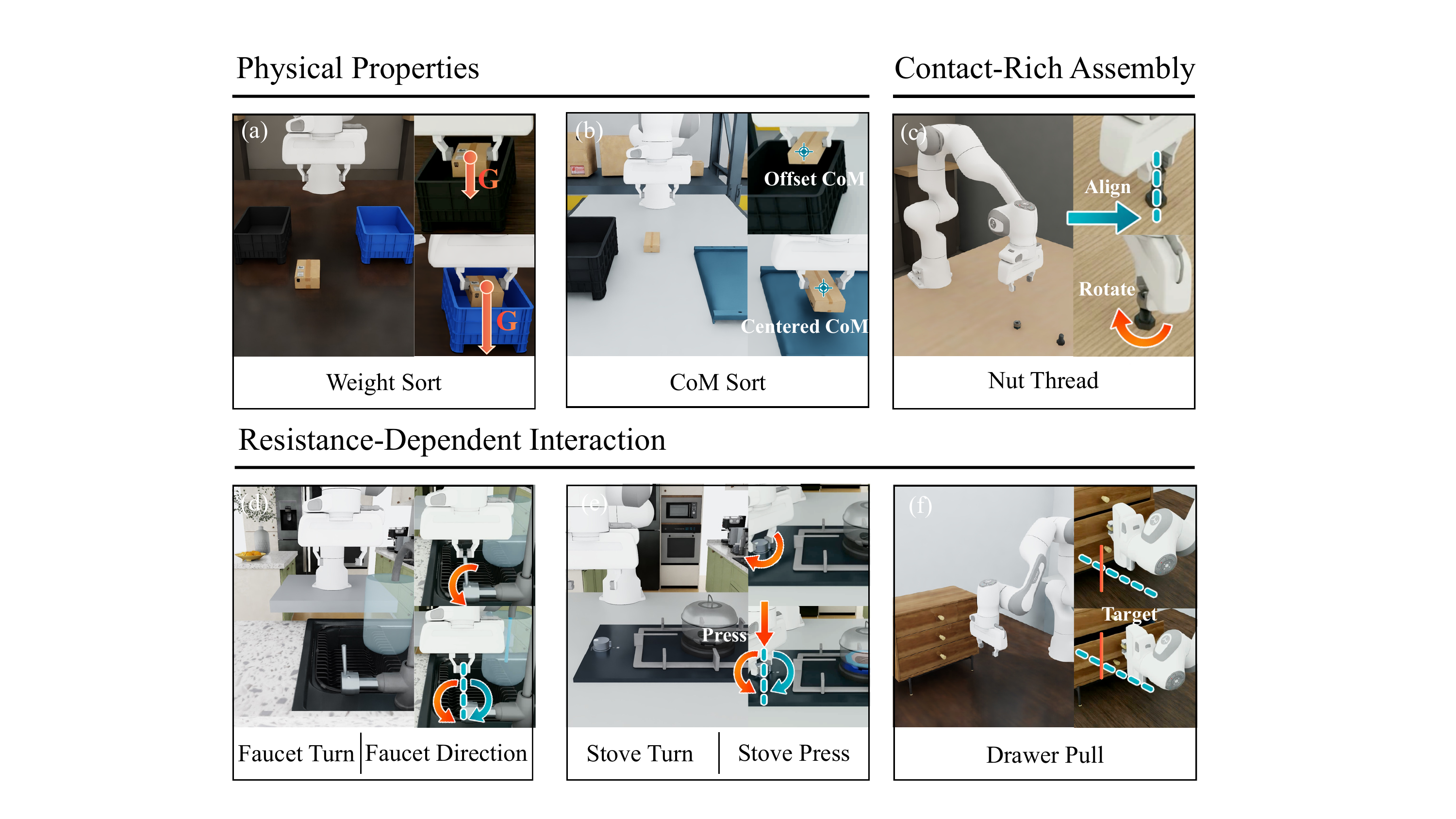}
\caption{Simulation tasks. (a) Weight Sort and (b) CoM Sort require
physical-property inference; (c) Nut Thread requires contact-rich
assembly; (d) Faucet Turn and Faucet Direction, (e) Stove Turn and
Stove Press, and (f) Drawer Pull require resistance-dependent
interaction. Insets mark the hidden property or resistance transition
that appearance does not reveal; Faucet Direction and Stove Press share
the faucet and stove environments, respectively.}
\label{fig:task_overview}
\end{figure}

\begin{figure*}[!t]
\centering
\includegraphics[width=\textwidth,trim=113bp 88bp 126bp 90bp,clip]{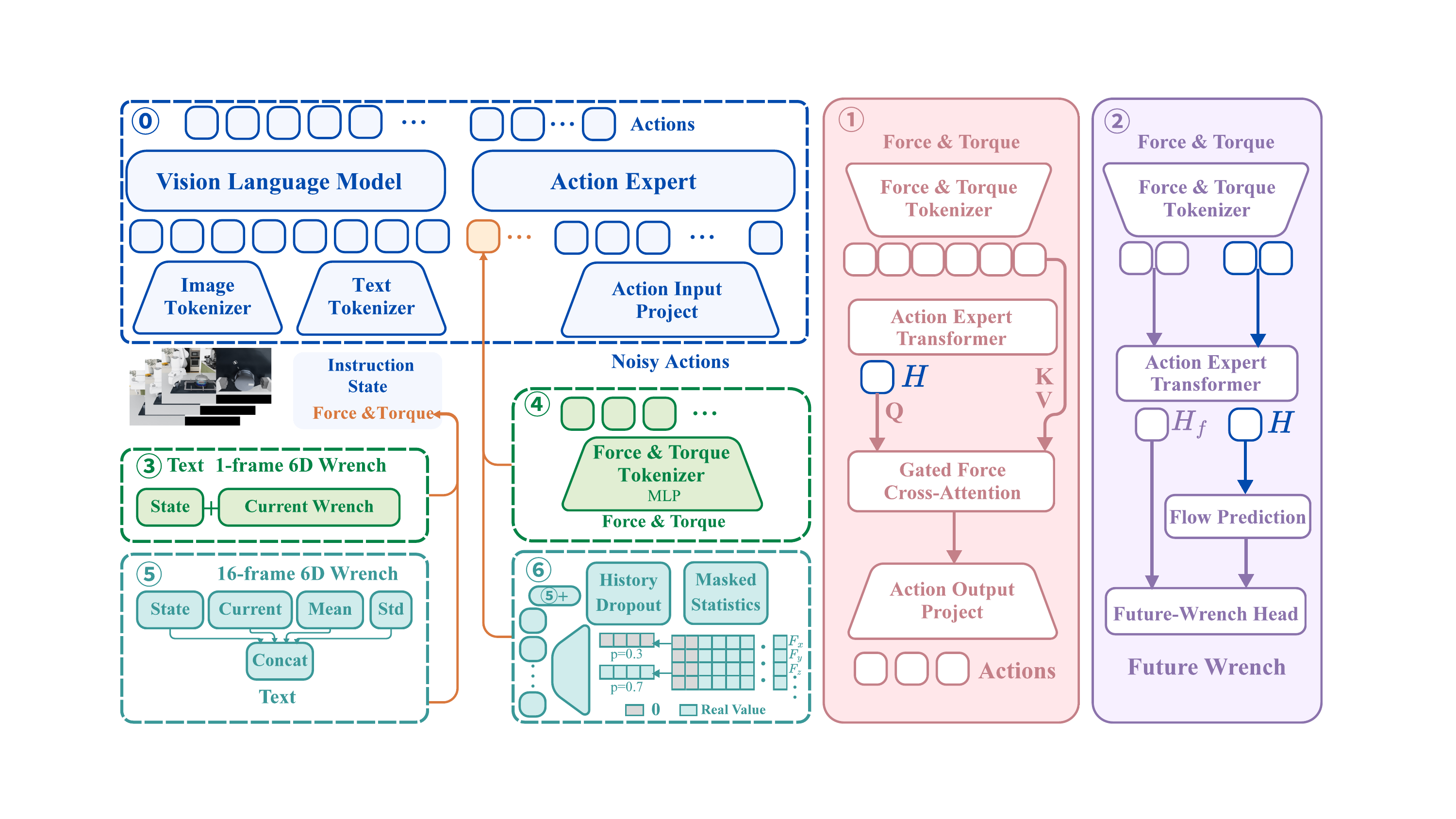}
\caption{Force integration on a shared \pifive{} backbone. \figcircled{1}--\figcircled{4} denote Cross-Attention, Token Fusion + Pred, Prompt Fusion, and Token Fusion, respectively; \pifive{} (No Force) denotes the baseline. \figcircled{5} and \figcircled{6} show the additional Temporal Prompt and Dual-Path Fusion references. Alternative pathways are compared separately. The future-force head in \figcircled{2} provides training-only supervision. Force and torque components together form the six-axis force observation.}
\label{fig:framework}
\end{figure*}

\section{ForceBench}
\label{sec:benchmark}
ForceBench evaluates six-axis wrist-force integration across
physical-property inference, resistance-dependent interaction, and
contact-rich assembly. Its workflow combines parameterized task instances,
synchronized expert demonstrations, and seeded policy evaluation under a
shared observation--action interface (Fig.~\ref{fig:pipeline}).
Task-specific decision and completion metrics distinguish physical
decisions from subsequent execution.

\subsection{Assets and Physical Mechanisms}
Built on Isaac Lab / Isaac Sim\cite{mittal2023orbit}, ForceBench uses a
Franka arm with a parallel gripper in tabletop and household environments.
Assets define geometry and articulation, while task parameters specify
mass, center of mass, resistance, and initial poses.
Resistance mechanisms couple force and moment profiles to articulation
state, producing direction-dependent resistance, rotational transitions,
and axial engagement before turning.
A shared task interface provides seeded initialization, interaction
dynamics, expert state access, and outcome evaluation.

\subsection{Task Design}
ForceBench comprises eight tasks across physical-property inference,
resistance-dependent interaction, and contact-rich assembly
(Fig.~\ref{fig:task_overview}). The first two categories vary physical
conditions without distinct visual appearances; assembly complements
them with thread engagement and constrained rotational insertion.

\emph{Physical-property inference} requires identifying an object
property through interaction and selecting the corresponding destination.
\begin{itemize}
\setlength{\itemsep}{2pt}\setlength{\parskip}{0pt}
\item \textbf{Weight Sort.}
Visually identical parcels have different masses. The robot lifts each
parcel and places it in the black container if light or the blue container
if heavy. Success requires correct placement, release, and settling.

\item \textbf{CoM Sort.}
Parcels share the same appearance and total mass but have centered or
offset centers of mass, producing different rotational loads about the
grasp. The robot holds each parcel steadily, then places centered parcels
on the conveyor and imbalanced parcels in the basket, with release and
settling required.
\end{itemize}

\emph{Resistance-dependent interaction} requires adapting direction,
engagement, or stopping to the load encountered during motion.
\begin{itemize}
\setlength{\itemsep}{2pt}\setlength{\parskip}{0pt}
\item \textbf{Faucet Turn.}
The robot opens a valve against resistance that peaks and then drops,
with the peak angle varying across instances. It must stop in the target
region just beyond the transition, release, and withdraw, leaving the
water on and the handle stationary.

\item \textbf{Faucet Direction.}
The low-resistance direction and peak location vary across instances.
Success requires crossing the resistance peak in the lower-resistance
direction and leaving the valve open, released, clear of the gripper,
and stationary. With direction sign $d\in\{-1,+1\}$, final valve
angle $\theta_f$, and peak-angle magnitude $C$, the final-angle criterion
is $d\theta_f\ge C$, with no upper-angle bound.
The gripper must be at least 65\% open, the TCP at least 9\,cm above
the closed-angle grasp height, and the absolute valve angular speed
at most $0.01$\,rad/s.

\item \textbf{Stove Turn.}
The robot grasps a stove knob and rotates through an ignition resistance
transition whose angular location varies across instances. Success
requires crossing the transition, reaching the prescribed terminal angle,
and activating the flame.

\item \textbf{Stove Press.}
A spring-loaded mechanism prevents rotation until the knob is sufficiently
depressed. The robot must coordinate axial pressing with rotation in the
low-resistance direction and cross the ignition transition. Direction and
peak location vary across instances; completion requires engagement and
ignition.

\item \textbf{Drawer Pull.}
The drawer encounters a rising-resistance region with varying onset,
extent, and breakaway resistance. The robot must stop at an opening
corresponding to a specified resistance level, then release with the
drawer stable. Pulling beyond the mechanism's limit counts as failure.
\end{itemize}

\emph{Contact-rich assembly} evaluates constrained execution commonly
studied by force-aware
policies\cite{yu2025forcevlaenhancingvlamodels,he2024foar}.
\begin{itemize}
\setlength{\itemsep}{2pt}\setlength{\parskip}{0pt}
\item \textbf{Nut Thread.}
The robot aligns a grasped nut with a bolt at a varying target
position, engages the threads, and rotates the nut to screw it down
to the prescribed threaded height, then releases and withdraws.
Success requires reaching the threaded target height while centered
and upright, together with release and gripper withdrawal.
The evaluator checks the resulting assembly geometry and completion
conditions; its numerical tolerances are given in
Appendix~\ref{app:simulation}.
\end{itemize}

Physical conditions may also affect visible motion and proprioception;
the comparison therefore measures the additional benefit of force.

\subsection{Expert Demonstrations and Dataset}
Task-specific scripted experts use privileged simulator state and a shared
controller to collect 1,600 successful demonstrations (200 per task).
Only trajectories satisfying the evaluation success criteria are retained;
privileged state is excluded from policy inputs.
Physics runs at 60\,Hz, while control and synchronized observation--action
recording operate at 15\,Hz. The policy observation is
\begin{equation}
o_t=(I_t^{\mathrm{ext}},I_t^{\mathrm{wrist}},s_t,\ell,f_t),
\end{equation}
where the external and wrist RGB images have resolution $224\times224$,
$s_t\in\mathbb{R}^7$ comprises tool center point (TCP) position, roll--pitch--yaw, and gripper
state, $\ell$ is the task instruction, and
$f_t=(F_x,F_y,F_z,T_x,T_y,T_z)\in\mathbb{R}^6$.
The wrench is the parent-on-child reaction at the wrist--hand joint,
expressed in the joint frame without gravity compensation.
Recorded actions specify absolute TCP poses and gripper targets.

\subsection{Parallel Evaluation and Outcomes}
Policies are evaluated in parallel on layouts initialized with shared
seeds, with action chunks executed through a common controller.
Success is scored per episode using the complete task-specific criteria,
including goal attainment and, where applicable, release, clearance,
stability, or latched completion events.
Per-episode records additionally retain destination choices, mechanism
progress, and assembly geometry for diagnostic use.
Appendices~\ref{app:simulation}--\ref{app:interpretation} give additional
task criteria, force-interface details, and metric definitions.

\section{Force Integration Methods}
\label{sec:methods}

To study how a policy turns force evidence into behavior, we separate two design choices: \emph{how force conditions action generation}
and \emph{whether predicting future force provides useful supervision}.
The observation comparison places the current force sample in the numerical
prompt, the action-token sequence, or an output cross-attention layer. The
predictive comparison then adds supervision while retaining the Token Fusion
input pathway. This separates the study of an interface from that of an
additional learning objective.
At query $t$, a shared \pifive{} backbone\cite{intelligence2025pi05visionlanguageactionmodelopenworld} maps $o_t$ to an action chunk
$A_t=(a_t,\ldots,a_{t+H-1})$, where $H$ is the action-chunk length.
The actions specify absolute TCP and gripper targets. Figure~\ref{fig:framework} shows the four primary force designs and the two temporal references. \pifive{} (No Force)
provides the baseline; the temporal systems provide additional reference points.
The input designs jointly vary representation, normalization, capacity, and
insertion location; the predictive variant keeps the Token Fusion input pathway
fixed. Temporal systems are separate references, not a controlled history-length
ablation.

\subsection{Force as a Policy Observation}
\textbf{\pifive{} (No Force).} This baseline fine-tunes the shared \pifive{} backbone on ForceBench demonstrations. It receives images, language, and proprioception, without force observations.
The four integration designs additionally receive the current six-axis force sample.

\textbf{Token Fusion.} A multilayer perceptron (MLP) maps force to six continuous
tokens placed before action tokens in the action-expert suffix. This provides
a continuous physical observation within action generation.

\textbf{Prompt Fusion.} Force is appended to robot state,
normalized, and discretized into numerical tokens in the prompt. This exposes the physical observation through the language input pathway.

\textbf{Cross-Attention.} Force tokens enter a cross-attention layer
between the action-expert output and action projection. This conditions the action representation near its output.

\subsection{Force as Predictive Supervision}
\textbf{Token Fusion + Pred.} This design uses Token Fusion and
adds ten-step future-force prediction during training. The head combines
contextual force tokens with a pooled projection of estimated clean actions.
It is not used to select or modify actions at deployment.
For a future horizon $K=10$, its training objective is
\begin{equation}
\mathcal{L}=\mathcal{L}_{\mathrm{flow}}+
\lambda\frac{\sum_{k=1}^{K}m_k\|\hat f_{t+k}-f_{t+k}\|_2^2}
{6\max(1,\sum_{k=1}^{K}m_k)},
\end{equation}
where $\mathcal{L}$ is the total training loss and
$\mathcal{L}_{\mathrm{flow}}$ is the backbone's action flow-matching loss.
The auxiliary weight is $\lambda=3\times10^{-4}$.
Here, $t$ denotes the current control step and $k$ indexes the next $K=10$
steps; $\hat f_{t+k}$ and $f_{t+k}\in\mathbb{R}^{6}$ are the predicted and
recorded six-axis force vectors in the training-normalized space.
The validity mask $m_k\in\{0,1\}$ is one for an available future target and
zero for padding. The squared Euclidean norm $\|\cdot\|_2^2$ sums errors
across the six components; the denominator averages over those components and
valid future steps, with $\max(1,\sum_k m_k)$ preventing division by zero.
Its pooled action context summarizes the chunk without retaining action order.

\subsection{Temporal Reference Systems}
Two separately trained policies extend the observation from a current force
sample to temporal statistics.

\textbf{Temporal Prompt.} The current-sample designs above use one force
measurement. Temporal Prompt instead reads the last 16 samples (about one second at
15\,Hz) and appends the current sample together with the per-axis mean and
standard deviation of the window to the robot state, which is then normalized
and discretized into numerical prompt tokens as in Prompt Fusion. The state vector
grows from 7 to 25 numbers; the backbone and the action expert are unchanged and
no parameter is added.

\textbf{Dual-Path Fusion.}
Dual-Path Fusion keeps the Temporal Prompt pathway and adds a second pathway into the action expert:
the same 16-sample window is reduced to 25 masked statistics (window mean and
standard deviation, means of the first and second half, and a presence flag),
mapped by a two-layer MLP to four continuous tokens placed before the action
tokens, with the Token Fusion attention layout. During training the window fed to this
pathway is zeroed for 30\% of the samples, so the policy also learns to act
from the prompt pathway alone; at deployment the window is always provided.
The training objective is the flow-matching loss only.

\section{Experiments and Analysis}
\label{sec:experiments}

\begin{table*}[t]
\centering
\caption{Simulation task success rates (\%), with 200 episodes per
task--method pair. Mean averages the eight tasks equally.}
\label{tab:preliminary}
\fontsize{8}{9.5}\selectfont
\renewcommand{\arraystretch}{1.12}
\setlength{\tabcolsep}{2pt}
\begin{tabular*}{\textwidth}{@{\extracolsep{\fill}}lccccccccc@{}}
\toprule
\multicolumn{1}{c}{Method}
& \shortstack{Weight\\Sort}
& \shortstack{CoM\\Sort}
& \shortstack{Faucet\\Turn}
& \shortstack{Faucet\\Direction}
& \shortstack{Stove\\Turn}
& \shortstack{Stove\\Press}
& \shortstack{Drawer\\Pull}
& \shortstack{Nut\\Thread}
& Mean $\uparrow$ \\
\midrule
\pifive{} (No Force)~\cite{intelligence2025pi05visionlanguageactionmodelopenworld}
& 54.0 & 45.5 & 9.5 & 31.5 & 8.5 & 1.0 & 4.5 & 31.5 & 23.25 \\
\phantom{\pifive{} (}\llap{$+$\,}Token Fusion
& 50.5 & 50.0 & 12.0 & 28.5 & 6.0 & 2.5 & 13.0 & 26.0 & 23.56 \\
\phantom{\pifive{} (}\llap{$+$\,}Prompt Fusion
& 88.0 & 56.5 & 21.0 & \textbf{41.5} & 16.0 & 2.5
& \textbf{23.0} & 34.0 & \textbf{35.31} \\
\phantom{\pifive{} (}\llap{$+$\,}Cross-Attention
& 54.5 & 51.0 & 6.5 & 33.0 & 5.5 & 2.5 & 6.5 & 17.5 & 22.13 \\
\phantom{\pifive{} (}\llap{$+$\,}Token Fusion + Pred
& 78.0 & 48.5 & 13.5 & 32.5 & 6.0 & 1.0 & 4.0 & 26.5 & 26.25 \\
\phantom{\pifive{} (}\llap{$+$\,}Temporal Prompt
& 90.0 & 69.5 & 17.5 & 31.0 & 7.0 & \textbf{21.5}
& 2.5 & 22.0 & 32.63 \\
\phantom{\pifive{} (}\llap{$+$\,}Dual-Path Fusion
& \textbf{99.5} & \textbf{72.0} & \textbf{22.0} & 30.0
& \textbf{17.5} & 19.0 & 4.5 & 16.5 & 35.13 \\
\midrule
$\pi_0$~\cite{black2024pi0}
& 61.5 & 39.0 & 3.5 & 27.0 & 7.0 & 5.0 & 7.5 & \textbf{36.0} & 23.31 \\
ForceVLA~\cite{yu2025forcevlaenhancingvlamodels}
& 35.5 & 45.5 & 12.5 & 14.0 & 10.0 & 12.0 & 20.0 & 26.0 & 21.94 \\
ForceFlow~\cite{forceflow2026}
& 51.5 & 3.0 & 0.0 & 0.5 & 0.0 & 0.0 & 0.0 & 0.0 & 6.88 \\
\bottomrule
\end{tabular*}
\par\vspace{2pt}
\begin{minipage}{\textwidth}
\fontsize{7}{8.2}\selectfont\raggedright
\end{minipage}
\end{table*}

\subsection{Experimental Design}

We compare seven \pifive{} configurations: a no-force baseline and
six force integration designs. All seven share the backbone and task
demonstrations and are trained for 10,000 steps with a batch size of 32.
Additional references include $\pi_0$~\cite{black2024pi0}
(10,000 training steps),
ForceVLA~\cite{yu2025forcevlaenhancingvlamodels}, and
ForceFlow~\cite{forceflow2026} (100,000 training steps).
For all methods, including the external references, a separate policy
is trained for each task using instances generated with seeds 0--199.

Each of the ten systems is evaluated over 200 episodes per task
across eight tasks, totaling 16,000 episodes.
Success follows Sec.~\ref{sec:benchmark}.
For Faucet Direction, all ten systems are evaluated using the same
lower-bound-only angle criterion.
Overall and category-level means weight their constituent tasks equally.
Task-level outcome and trajectory diagnostics complement success rates
by examining physical decisions and execution requirements.
A real-world study evaluates eight systems on five tasks
(Sec.~\ref{sec:realworld}).

\textbf{Simulation execution protocol.}
All methods use the same evaluation seeds (300--499) for each task.
Physics runs at 60\,Hz and control at 15\,Hz, with a budget of
1,500 control steps (100\,s of simulated time) per episode.
Actions specify absolute TCP poses and gripper targets, executed
by a damped-least-squares inverse-kinematics controller with
synchronized quintic joint interpolation.
The seven \pifive{} configurations, $\pi_0$, and ForceVLA predict
50-step action chunks and execute all 50 steps before querying the
policy with updated observations, unless the episode terminates earlier.
ForceFlow predicts 64 steps and
executes the first 32 before the next query.
Thus, 15\,Hz denotes the control rate, not the policy-query rate;
the external references retain this difference in execution horizon.

\subsection{Overall Performance}

Prompt Fusion achieves the highest observed mean simulation success
of 35.31\%, exceeding the no-force baseline by 12.06 percentage points
(Table~\ref{tab:preliminary}), followed closely by Dual-Path Fusion
at 35.13\%.

Dual-Path Fusion leads physical-property inference (85.75\%),
while Prompt Fusion leads resistance-dependent interaction (20.80\%).
On Nut Thread, Prompt Fusion ranks highest among the seven
\pifive{} configurations (34.00\%), although the $\pi_0$ reference
reaches 36.00\%.

Future-force supervision raises Token Fusion's overall mean
from 23.56\% to 26.25\%.
Its physical-property inference gains are driven by Weight Sort,
while CoM Sort success decreases, indicating task-dependent benefits.

\begin{figure*}[!t]
\centering
\includegraphics[width=\textwidth]{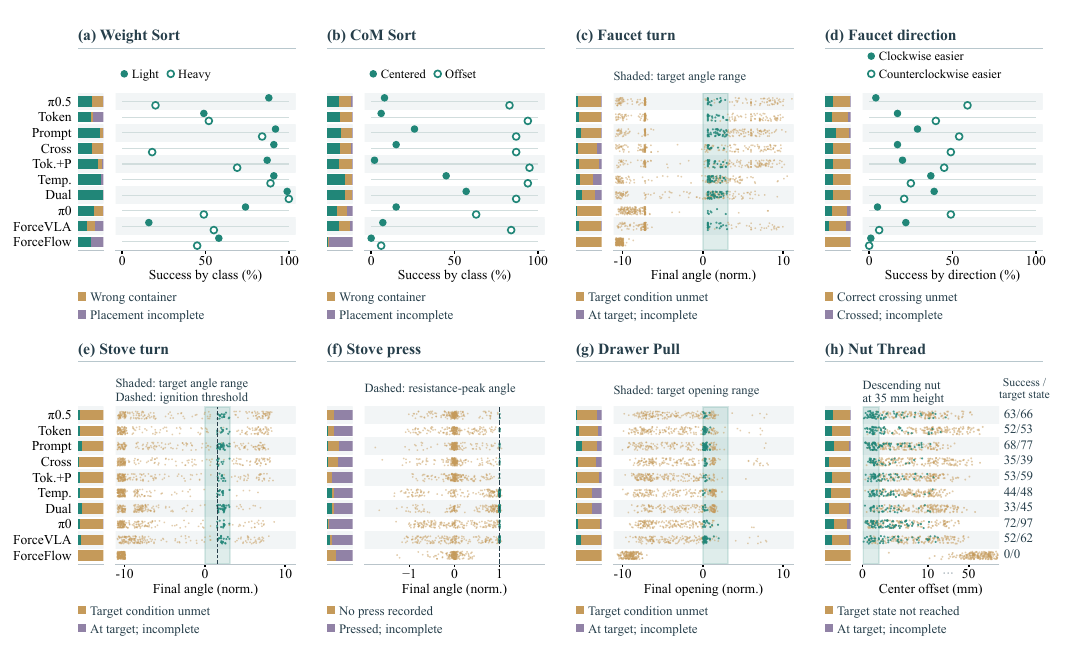}
\caption{Task outcomes and state diagnostics
(200 episodes per task--method pair).
Bars show success (green) and labeled failure categories
(ochre/purple).
In (a,b,d), filled/open markers report success rates for the
labeled physical conditions.
In (c,e,f,g,h), dots represent episode measurements colored
by success (green) or failure (ochre).
In (c,e,g), shaded target intervals map to $[0,1]$;
reaching these intervals alone does not establish full success.
In (f), angles are normalized by the direction-specific
resistance-peak angle.
In (h), offset is measured during descent at a nut height of
35~mm on a symmetric-log axis; the ellipsis omits tick labels only.
Count pairs report full-task successes / episodes reaching
the target assembly state (centered, upright, and at the target
height); full success additionally requires release and withdrawal.}

\label{fig:category_results}
\end{figure*}

\begin{table*}[!t]
\centering
\caption{Real-world task success rates (\%), with 20 trials per task--method pair.
The mean weights all five tasks equally; bold denotes the highest rate in each column.}
\label{tab:hardware_results}
\fontsize{8}{9.5}\selectfont
\setlength{\tabcolsep}{3pt}
\renewcommand{\arraystretch}{1.12}
\begin{tabular*}{\textwidth}{@{\extracolsep{\fill}}lcccccc@{}}
\toprule
& \multicolumn{1}{c}{Property inference}
& \multicolumn{3}{c}{Resistance-dependent interaction}
& \multicolumn{1}{c}{Assembly} & \\
\cmidrule(lr){2-2}\cmidrule(lr){3-5}\cmidrule(lr){6-6}
Method & Weight Sort & Plate Retrieve & Stamp Press & Plug Insert & Block Insert & Mean $\uparrow$ \\
\midrule
\pifive{} (No Force) & 30.0 & 20.0 & 40.0 & 60.0 & 50.0 & 40.00 \\
Token Fusion & 45.0 & 40.0 & 50.0 & 60.0 & 10.0 & 41.00 \\
Prompt Fusion & 70.0 & 40.0 & 65.0 & 80.0 & \textbf{60.0} & \textbf{63.00} \\
Cross-Attention & 60.0 & 30.0 & 40.0 & 50.0 & 55.0 & 47.00 \\
Token Fusion + Pred & 45.0 & \textbf{60.0} & 60.0 & 45.0 & 35.0 & 49.00 \\
Temporal Prompt & 80.0 & 45.0 & 65.0 & 75.0 & 35.0 & 60.00 \\
Dual-Path Fusion & \textbf{95.0} & 45.0 & \textbf{70.0} & \textbf{85.0} & 20.0 & \textbf{63.00} \\
\midrule
ForceVLA & 50.0 & 0.0 & 40.0 & 70.0 & 5.0 & 33.00 \\
\bottomrule
\end{tabular*}
\end{table*}

\subsection{Analysis and Discussion}

\textbf{Force encoding should match the backbone's conditioning pathway.}
Prompt Fusion appends force to proprioceptive state and reuses
\pifive{}'s existing numerical-conditioning pathway.
Since wrist loads depend on both object properties and the current
interaction, this joint input provides relevant context for interpreting
physical feedback. Reusing a pretrained conditioning format may also
reduce the adaptation required to incorporate force under a fixed
training budget. The gains on physical-property and resistance-dependent
tasks are consistent with this interpretation, suggesting that force
encoding should be designed jointly with the backbone's conditioning
mechanism.

\textbf{Temporal statistics match the persistence of physical properties.}
Mass and center of mass remain fixed within an episode, whereas
instantaneous wrist measurements vary with motion and contact.
The temporal designs combine current measurements with window
statistics that summarize sustained loads and their variability.
These statistics aggregate repeated observations associated with the
same underlying property. Their stronger performance on property
tasks is consistent with persistent physical cues becoming more
accessible to the policy, suggesting that temporal inputs should
reflect the time scale of task-relevant information.

\textbf{Future-force supervision links action representations to physical response.}
The prediction head combines contextual force tokens with pooled
features of estimated clean actions to predict subsequent interaction
loads. This objective provides supervision on the physical consequences
of estimated motion beyond matching demonstrated actions, potentially
encouraging representations useful for execution.
The Weight Sort improvement is consistent with this role.
Because the head is used only during training, future physical feedback
serves as a learning target without introducing an additional predictive
controller at deployment.

\subsection{What the Failure Structure Reveals}

Figure~\ref{fig:category_results} distinguishes task-specific physical
outcomes from the additional requirements for complete success.

\textbf{Effective force integration improves choices governed by hidden physical conditions.}
For Prompt Fusion, the Weight Sort gain over the no-force baseline
is concentrated in destination outcomes: wrong destinations decrease
from 86 to 17 episodes, while incomplete placements remain nearly
unchanged (6 versus 7).
On Faucet Turn, Prompt Fusion satisfies the target-angle and crossing
conditions in 45 episodes, compared with 26 for the baseline, while
complete successes increase from 19 to 42.
These results associate the gains with more appropriate destination
and stopping outcomes, supporting the use of force to guide behavior
under hidden physical properties and resistance transitions.

\textbf{Predictive and temporal gains follow different decision--execution tradeoffs.}
Adding predictive supervision to Token Fusion reduces incomplete
Weight Sort placements from 84 to 14 and raises successes from
101 to 156, while wrong destinations increase from 15 to 30.
The net gain is therefore dominated by placement completion.
Because more episodes reach a destination, the wrong-destination
counts alone do not establish lower decision accuracy.

Temporal designs exhibit a different pattern.
Temporal Prompt satisfies the Faucet Turn angle-and-crossing
conditions more often than Prompt Fusion (98 versus 45 episodes),
yet succeeds less often (35 versus 42).
On Drawer Pull, Dual-Path Fusion and Prompt Fusion both satisfy
the opening-range and intact-latch conditions in 81 episodes,
but achieve 9 and 46 successes, respectively.
Thus, greater or unchanged satisfaction of selected object-state
conditions can coexist with lower complete-task success.
These differences show that richer force representations and
objectives can redistribute gains and failures across task
requirements rather than improve them uniformly.
Interface comparisons should therefore consider the distribution
of outcomes alongside aggregate success.

\begin{figure}[!t]
\centering
\includegraphics[
    width=\columnwidth,
    trim=15bp 245bp 17bp 194bp,
    clip
]{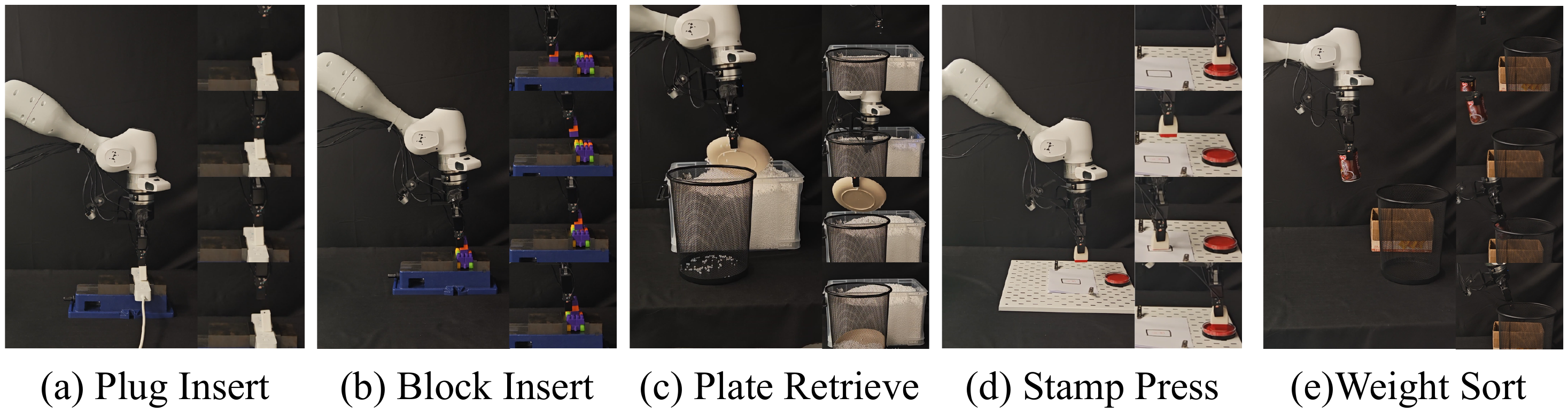}
\caption{Real-world tasks. (a) Plug Insert, (b) Block Insert,
(c) Plate Retrieve, (d) Stamp Press, and (e) Weight Sort.
Each panel shows key interaction frames from the external camera.}
\label{fig:real_tasks}

\vspace{6pt}
\includegraphics[
    width=\columnwidth,
    trim=65bp 104bp 50bp 20bp,
    clip
]{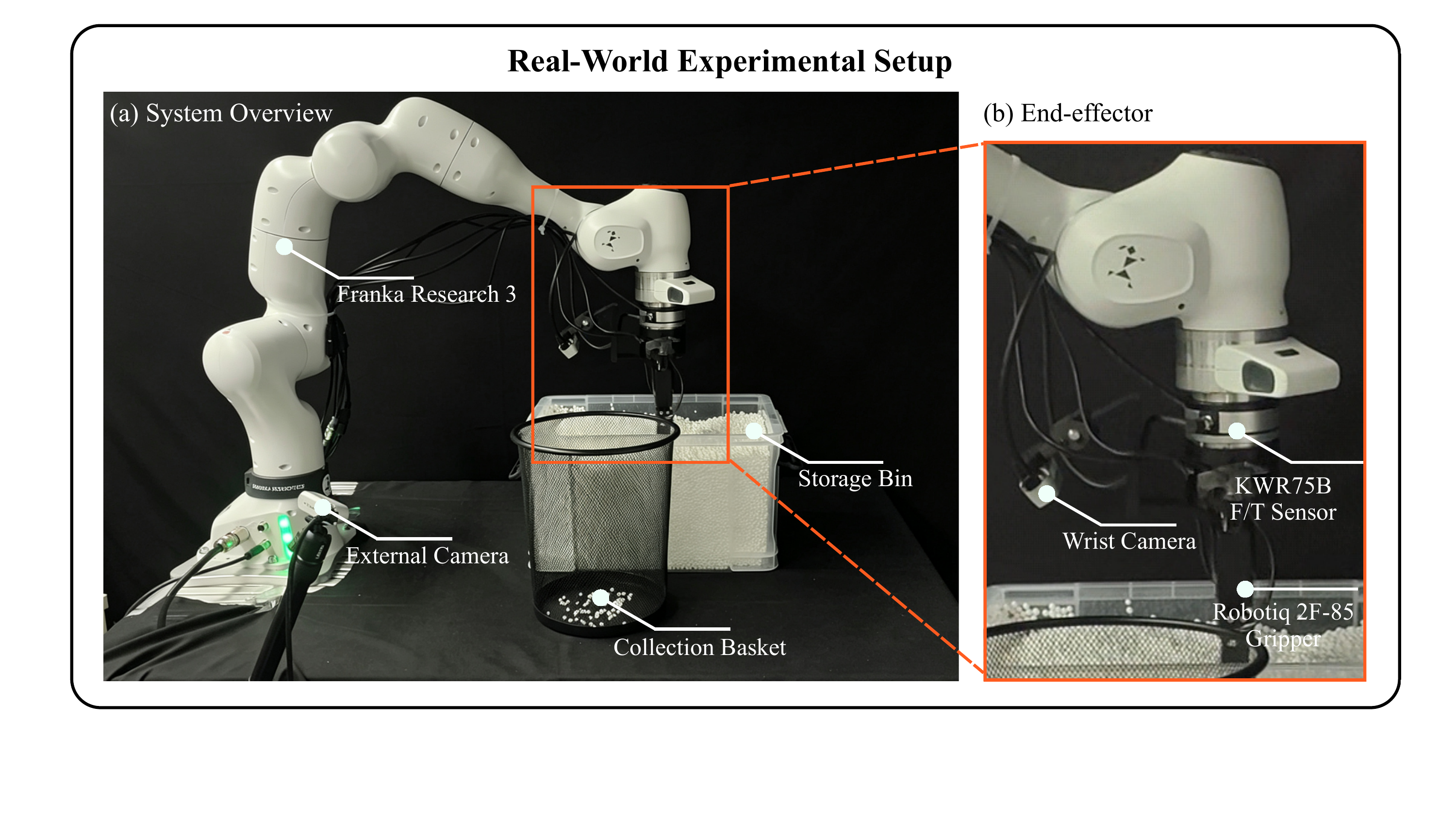}
\caption{Real-world setup. (a) Franka Research 3 arm with the external
camera and collection basket; (b) end-effector
with the wrist camera, KWR75B force/torque sensor, and Robotiq gripper.}
\label{fig:real_setup}
\end{figure}

\subsection{Alignment with Real-World Manipulation}
\label{sec:realworld}

We evaluate eight systems on five real-world tasks covering the same
three manipulation categories as ForceBench
(Fig.~\ref{fig:real_tasks}).
Experiments use a Franka Research 3 equipped with a six-axis
force/torque sensor (Fig.~\ref{fig:real_setup}).
Each system uses 200 demonstrations per task and is trained for
10,000 steps with a batch size of 64.
Evaluation uses the same 20 initial conditions per task across
methods, totaling 800 trials.

The real-world results reproduce the main qualitative trends observed
in simulation (Table~\ref{tab:hardware_results}).
Prompt Fusion and Dual-Path Fusion achieve the highest observed mean
success of 63.00\%, compared with 40.00\% for the no-force baseline.
Prompt Fusion improves success on every task, while both temporal
designs exceed Prompt Fusion on Weight Sort but underperform the
baseline on Block Insert.
This agreement supports the practical relevance of \bench{}:
the benchmark captures task-dependent benefits and limitations
of force integration that are also observed in physical manipulation.

\section{Conclusion}
\label{sec:conclusion}

We introduced \bench{}, a simulation benchmark for evaluating six-axis
wrist-force integration in VLA policies. It combines eight tasks,
1,600 synchronized demonstrations, and unified evaluation across
physical-property inference, resistance-dependent interaction,
and contact-rich assembly.

On a shared \pifive{} backbone, Prompt Fusion achieves the highest
observed mean simulation success and the most consistent task-wise
improvements. In simulation, predictive supervision and temporal
representations primarily benefit physical-property tasks, while
both temporal designs underperform the no-force baseline on assembly.
Real-world evaluation supports these qualitative trends:
Prompt Fusion and Dual-Path Fusion tie at 63.00\% mean success,
compared with 40.00\% for the baseline, but only Prompt Fusion
improves all five tasks. These findings support \bench{}'s practical
value for identifying the task-dependent benefits and limitations
of force integration.

\bibliographystyle{IEEEtran}
\bibliography{refs}

\begin{thebibliography}{10}
\providecommand{\url}[1]{#1}
\csname url@samestyle\endcsname
\providecommand{\newblock}{\relax}
\providecommand{\bibinfo}[2]{#2}
\providecommand{\BIBentrySTDinterwordspacing}{\spaceskip=0pt\relax}
\providecommand{\BIBentryALTinterwordstretchfactor}{4}
\providecommand{\BIBentryALTinterwordspacing}{\spaceskip=\fontdimen2\font plus
\BIBentryALTinterwordstretchfactor\fontdimen3\font minus
  \fontdimen4\font\relax}
\providecommand{\BIBforeignlanguage}[2]{{%
\expandafter\ifx\csname l@#1\endcsname\relax
\typeout{** WARNING: IEEEtran.bst: No hyphenation pattern has been}%
\typeout{** loaded for the language `#1'. Using the pattern for}%
\typeout{** the default language instead.}%
\else
\language=\csname l@#1\endcsname
\fi
#2}}
\providecommand{\BIBdecl}{\relax}
\BIBdecl

\bibitem{kim2024openvla}
\BIBentryALTinterwordspacing
M.~J. Kim, K.~Pertsch, S.~Karamcheti \emph{et~al.}, ``{OpenVLA: An Open-Source
  Vision-Language-Action Model},'' 2024. [Online]. Available:
  \url{https://arxiv.org/abs/2406.09246}
\BIBentrySTDinterwordspacing

\bibitem{black2024pi0}
\BIBentryALTinterwordspacing
K.~Black, N.~Brown, D.~Driess \emph{et~al.}, ``{$\pi_0$: A
  Vision-Language-Action Flow Model for General Robot Control},'' 2024.
  [Online]. Available: \url{https://arxiv.org/abs/2410.24164}
\BIBentrySTDinterwordspacing

\bibitem{intelligence2025pi05visionlanguageactionmodelopenworld}
\BIBentryALTinterwordspacing
{Physical Intelligence}, K.~Black, N.~Brown \emph{et~al.}, ``{$\pi_{0.5}$: a
  Vision-Language-Action Model with Open-World Generalization},'' 2025.
  [Online]. Available: \url{https://arxiv.org/abs/2504.16054}
\BIBentrySTDinterwordspacing

\bibitem{kim2025oft}
M.~J. Kim, C.~Finn, and P.~Liang, ``Fine-tuning vision-language-action models:
  Optimizing speed and success,'' in \emph{Robotics: Science and Systems},
  2025, arXiv:2502.19645.

\bibitem{chi2023diffusion}
\BIBentryALTinterwordspacing
C.~Chi, Z.~Xu, S.~Feng \emph{et~al.}, ``{Diffusion Policy: Visuomotor Policy
  Learning via Action Diffusion},'' 2023. [Online]. Available:
  \url{https://arxiv.org/abs/2303.04137}
\BIBentrySTDinterwordspacing

\bibitem{hou2024acp}
\BIBentryALTinterwordspacing
Y.~Hou, Z.~Liu, C.~Chi \emph{et~al.}, ``{Adaptive Compliance Policy: Learning
  Approximate Compliance for Diffusion Guided Control},'' 2024. [Online].
  Available: \url{https://arxiv.org/abs/2410.09309}
\BIBentrySTDinterwordspacing

\bibitem{liu2025factr}
\BIBentryALTinterwordspacing
J.~J. Liu, Y.~Li, K.~Shaw, T.~Tao, R.~Salakhutdinov, and D.~Pathak, ``{FACTR:
  Force-Attending Curriculum Training for Contact-Rich Policy Learning},''
  2025. [Online]. Available: \url{https://arxiv.org/abs/2502.17432}
\BIBentrySTDinterwordspacing

\bibitem{xue2025rdp}
\BIBentryALTinterwordspacing
H.~Xue, J.~Ren, W.~Chen \emph{et~al.}, ``{Reactive Diffusion Policy: Slow-Fast
  Visual-Tactile Policy Learning for Contact-Rich Manipulation},'' 2025.
  [Online]. Available: \url{https://arxiv.org/abs/2503.02881}
\BIBentrySTDinterwordspacing

\bibitem{yu2025forcevlaenhancingvlamodels}
\BIBentryALTinterwordspacing
J.~Yu, H.~Liu, Q.~Yu \emph{et~al.}, ``{ForceVLA: Enhancing VLA Models with a
  Force-aware MoE for Contact-rich Manipulation},'' 2025. [Online]. Available:
  \url{https://arxiv.org/abs/2505.22159}
\BIBentrySTDinterwordspacing

\bibitem{li2026favla}
\BIBentryALTinterwordspacing
Y.~Li, P.~Tang, W.~Zhang \emph{et~al.}, ``{FAVLA: A Force-Adaptive Fast-Slow
  VLA model for Contact-Rich Robotic Manipulation},'' 2026. [Online].
  Available: \url{https://arxiv.org/abs/2602.23648}
\BIBentrySTDinterwordspacing

\bibitem{wang2026phaforce}
\BIBentryALTinterwordspacing
M.~Wang, Z.~Yue, R.~Lu \emph{et~al.}, ``{PhaForce: Phase-Scheduled Visual-Force
  Policy Learning with Slow Planning and Fast Correction for Contact-Rich
  Manipulation},'' 2026. [Online]. Available:
  \url{https://arxiv.org/abs/2603.08342}
\BIBentrySTDinterwordspacing

\bibitem{he2024foar}
\BIBentryALTinterwordspacing
Z.~He, H.~Fang, J.~Chen, H.-S. Fang, and C.~Lu, ``{FoAR: Force-Aware Reactive
  Policy for Contact-Rich Robotic Manipulation},'' 2024. [Online]. Available:
  \url{https://arxiv.org/abs/2411.15753}
\BIBentrySTDinterwordspacing

\bibitem{deng2026facet}
\BIBentryALTinterwordspacing
H.~Deng, H.~Liu, W.~Guo \emph{et~al.}, ``{Facet-0: A Robotic Foundation Model
  for Contact-Rich Precise Manipulation},'' 2026. [Online]. Available:
  \url{https://arxiv.org/abs/2609.01596}
\BIBentrySTDinterwordspacing

\bibitem{zhao2026fdvla}
\BIBentryALTinterwordspacing
R.~Zhao, W.~Wang, Y.~Ma \emph{et~al.}, ``{FD-VLA: Force-Distilled
  Vision-Language-Action Model for Contact-Rich Manipulation},'' 2026.
  [Online]. Available: \url{https://arxiv.org/abs/2602.02142}
\BIBentrySTDinterwordspacing

\bibitem{fang2026domino}
H.~Fang, S.~Li, S.~Wang, X.~Xi, D.~Liang, and X.~Bai, ``Towards generalizable
  robotic manipulation in dynamic environments,'' in \emph{European Conference
  on Computer Vision (ECCV)}, 2026, arXiv:2603.15620.

\bibitem{lan2025autobio}
\BIBentryALTinterwordspacing
Z.~Lan, Y.~Jiang, R.~Wang \emph{et~al.}, ``{AutoBio: A Simulation and Benchmark
  for Robotic Automation in Digital Biology Laboratory},'' 2025. [Online].
  Available: \url{https://arxiv.org/abs/2505.14030}
\BIBentrySTDinterwordspacing

\bibitem{huang2025tactilevla}
\BIBentryALTinterwordspacing
J.~Huang, S.~Wang, F.~Lin, Y.~Hu, C.~Wen, and Y.~Gao, ``{Tactile-VLA: Unlocking
  Vision-Language-Action Model's Physical Knowledge for Tactile
  Generalization},'' 2025. [Online]. Available:
  \url{https://arxiv.org/abs/2507.09160}
\BIBentrySTDinterwordspacing

\bibitem{li2026forcevla2}
\BIBentryALTinterwordspacing
Y.~Li, Zhaxizhuoma, H.~Jiang \emph{et~al.}, ``{ForceVLA2: Unleashing Hybrid
  Force-Position Control with Force Awareness for Contact-Rich Manipulation},''
  2026. [Online]. Available: \url{https://arxiv.org/abs/2603.15169}
\BIBentrySTDinterwordspacing

\bibitem{lee2025manipforce}
\BIBentryALTinterwordspacing
G.~Lee, Y.~Lee, K.~Kim \emph{et~al.}, ``{ManipForce: Force-Guided Policy
  Learning with Frequency-Aware Representation for Contact-Rich
  Manipulation},'' 2025. [Online]. Available:
  \url{https://arxiv.org/abs/2509.19047}
\BIBentrySTDinterwordspacing

\bibitem{zhang2025tavla}
\BIBentryALTinterwordspacing
Z.~Zhang, H.~Xu, Z.~Yang \emph{et~al.}, ``{TA-VLA: Elucidating the Design Space
  of Torque-aware Vision-Language-Action Models},'' 2025. [Online]. Available:
  \url{https://arxiv.org/abs/2509.07962}
\BIBentrySTDinterwordspacing

\bibitem{forceflow2026}
\BIBentryALTinterwordspacing
S.~Zhang, Y.~Yuan, H.~Tang \emph{et~al.}, ``{ForceFlow: Learning to Feel and
  Act via Contact-Driven Flow Matching},'' 2026. [Online]. Available:
  \url{https://arxiv.org/abs/2605.11048}
\BIBentrySTDinterwordspacing

\bibitem{lee2026moss}
\BIBentryALTinterwordspacing
J.~Lee, H.~Jang, M.~Koo, J.~Park, and J.~Shin, ``{Modular Sensory Stream for
  Integrating Physical Feedback in Vision-Language-Action Models},'' 2026.
  [Online]. Available: \url{https://arxiv.org/abs/2604.23272}
\BIBentrySTDinterwordspacing

\bibitem{he2026fawam}
\BIBentryALTinterwordspacing
H.~He, Z.~Yan, Q.~Liu, N.~Guo, and W.~Lian, ``{FAWAM: Force-Aware World Action
  Models for Closed-Loop Contact-Rich Manipulation},'' 2026. [Online].
  Available: \url{https://arxiv.org/abs/2606.08555}
\BIBentrySTDinterwordspacing

\bibitem{liu2024forcemimic}
\BIBentryALTinterwordspacing
W.~Liu, J.~Wang, Y.~Wang, W.~Wang, and C.~Lu, ``{ForceMimic: Force-Centric
  Imitation Learning with Force-Motion Capture System for Contact-Rich
  Manipulation},'' 2024. [Online]. Available:
  \url{https://arxiv.org/abs/2410.07554}
\BIBentrySTDinterwordspacing

\bibitem{lou2026dreamtac}
\BIBentryALTinterwordspacing
Y.~Lou, Y.~Ye, Y.~Fu \emph{et~al.}, ``{Dream-Tac: A Unified Tactile World
  Action Model for Contact-Rich Robot Manipulation},'' 2026. [Online].
  Available: \url{https://arxiv.org/abs/2606.08737}
\BIBentrySTDinterwordspacing

\bibitem{zhang2026contactworld}
\BIBentryALTinterwordspacing
Z.~Zhang, P.~Zhou, K.~Zhang \emph{et~al.}, ``{ContactWorld: What
  Representations Matter in Vision-Tactile World Models for Contact-Rich
  Manipulation},'' 2026. [Online]. Available:
  \url{https://arxiv.org/abs/2606.13877}
\BIBentrySTDinterwordspacing

\bibitem{james2019rlbenchrobotlearningbenchmark}
\BIBentryALTinterwordspacing
S.~James, Z.~Ma, D.~R. Arrojo, and A.~J. Davison, ``{RLBench: The Robot
  Learning Benchmark \& Learning Environment},'' 2019. [Online]. Available:
  \url{https://arxiv.org/abs/1909.12271}
\BIBentrySTDinterwordspacing

\bibitem{yu2021metaworldbenchmarkevaluationmultitask}
\BIBentryALTinterwordspacing
T.~Yu, D.~Quillen, Z.~He \emph{et~al.}, ``{Meta-World: A Benchmark and
  Evaluation for Multi-Task and Meta Reinforcement Learning},'' 2019. [Online].
  Available: \url{https://arxiv.org/abs/1910.10897}
\BIBentrySTDinterwordspacing

\bibitem{mees2022calvin}
O.~Mees, L.~Hermann, E.~Rosete-Beas, and W.~Burgard, ``{CALVIN}: A benchmark
  for language-conditioned policy learning for long-horizon robot manipulation
  tasks,'' \emph{IEEE Robotics and Automation Letters}, 2022, arXiv:2112.03227.

\bibitem{liu2023libero}
B.~Liu, Y.~Zhu, C.~Gao \emph{et~al.}, ``{LIBERO}: Benchmarking knowledge
  transfer for lifelong robot learn\-ing,'' in \emph{Advances in Neural
  Information Processing Systems (NeurIPS), Datasets and Benchmarks Track},
  2023, arXiv:2306.03310.

\bibitem{nasiriany2024robocasalargescalesimulationeveryday}
\BIBentryALTinterwordspacing
S.~Nasiriany, A.~Maddukuri, L.~Zhang \emph{et~al.}, ``{RoboCasa: Large-Scale
  Simulation of Everyday Tasks for Generalist Robots},'' 2024. [Online].
  Available: \url{https://arxiv.org/abs/2406.02523}
\BIBentrySTDinterwordspacing

\bibitem{chen2025robotwin20scalabledata}
\BIBentryALTinterwordspacing
T.~Chen, Z.~Chen, B.~Chen \emph{et~al.}, ``{RoboTwin 2.0: A Scalable Data
  Generator and Benchmark with Strong Domain Randomization for Robust Bimanual
  Robotic Manipulation},'' 2025. [Online]. Available:
  \url{https://arxiv.org/abs/2506.18088}
\BIBentrySTDinterwordspacing

\bibitem{li2024evaluatingrealworldrobotmanipulation}
\BIBentryALTinterwordspacing
X.~Li, K.~Hsu, J.~Gu \emph{et~al.}, ``{Evaluating Real-World Robot Manipulation
  Policies in Simulation},'' 2024. [Online]. Available:
  \url{https://arxiv.org/abs/2405.05941}
\BIBentrySTDinterwordspacing

\bibitem{chen2026univtac}
\BIBentryALTinterwordspacing
B.~Chen, W.~Wan, T.~Chen \emph{et~al.}, ``{UniVTAC: A Unified Simulation
  Platform for Visuo-Tactile Manipulation Data Generation, Learning, and
  Benchmarking},'' 2026. [Online]. Available:
  \url{https://arxiv.org/abs/2602.10093}
\BIBentrySTDinterwordspacing

\bibitem{mittal2023orbit}
M.~Mittal, C.~Yu, Q.~Yu \emph{et~al.}, ``Orbit: A unified simulation framework
  for interactive robot learning environments,'' \emph{IEEE Robotics and
  Automation Letters}, vol.~8, no.~6, pp. 3740--3747, 2023, the framework is
  now distributed as NVIDIA Isaac Lab.

\end{thebibliography}
\clearpage
\useRomanappendicesfalse
\appendices
\section{Simulation Protocol and Task Criteria}
\label{app:simulation}

This appendix supplements the task definitions in
Sec.~\ref{sec:benchmark}. A task instance fixes the physical mechanism
and initial configuration through an environment seed. The training
instances use seeds 0--199, and evaluation uses the shared seeds
300--499, with 200 episodes per task and method. These are environment
seeds, not independent training runs. Each task has a separately trained
policy and 200 successful expert demonstrations. The same demonstration
set is used by the methods compared on that task.

\subsection{Observation, Action, and Execution}
The two RGB observations are an external view and a wrist view, each
resized to $224\times224$. The seven robot-state components are TCP
position, roll--pitch--yaw, and gripper state. The six force components
are ordered as $(F_x,F_y,F_z,T_x,T_y,T_z)$: force in newtons and torque
in newton-metres. The simulator supplies the parent-on-child reaction
wrench at the wrist--hand joint, expressed in that joint frame without
gravity compensation. Privileged mechanism parameters used by the
scripted experts are excluded from the policy observation.

Actions specify absolute TCP poses and gripper targets. A
damped-least-squares inverse-kinematics controller and synchronized
quintic joint interpolation execute these targets. Physics advances at
60\,Hz; action execution and recorded observations advance at 15\,Hz.
The episode budget is 1,500 control steps, equivalent to 100\,s of
simulated time.

The seven \pifive{} systems, $\pi_0$, and ForceVLA predict and execute
50 actions per query. ForceFlow predicts 64 actions and executes 32.
For an uninterrupted chunk, these execution intervals correspond to
$50/15\approx3.33$\,s and $32/15\approx2.13$\,s of simulated time,
respectively. These values describe observation refresh during chunked
execution, not measured inference latency. Episodes may terminate
before a complete chunk is executed.

\subsection{Physical Parameters and Completion Conditions}
Gripper-opening thresholds below use normalized opening, where one
denotes fully open. Unless a condition is explicitly described as a
latched event, entering an intermediate target region does not replace
the complete episode-level outcome. A latched event records that the
required event occurred earlier in the trajectory.

\paragraph{Weight Sort and CoM Sort}
Weight Sort varies parcel mass while keeping visual appearance fixed;
CoM Sort varies the center of mass while keeping appearance and total
mass fixed. Their decision component is selection of the appropriate
destination. Their completion component additionally requires placement,
release, and settling. A correct destination choice alone is therefore
insufficient for success.

\paragraph{Faucet Turn}
The resistance-peak angle $C$ is sampled uniformly from
$20^\circ$ to $30^\circ$. Completion requires detent crossing and
a final angle in $[C,C+2.1^\circ]$, with water visible. The gripper
must be at least 65\% open, the TCP at least 90\,mm above the
closed-angle grasp height, and the absolute valve angular speed at
most $0.01$\,rad/s. The target angle, release, clearance, and stationary
conditions are checked together.

\paragraph{Faucet Direction}
The peak magnitude is also sampled uniformly from $20^\circ$ to
$30^\circ$, and a seeded hidden direction determines the easier turn.
For direction sign $d$ and final angle $\theta_f$, the reported score
uses $d\theta_f\ge C$ without an upper angle bound, together with
detent crossing, water visibility, and the same release, clearance,
and angular-speed thresholds as Faucet Turn. This is a criterion on
the recorded final state: reaching the angle at any earlier frame
does not by itself establish complete success. Applying this scoring
rule to stored trajectories does not change their executed actions
or historical stopping times.

\paragraph{Stove Turn}
The peak angle $C$ is sampled uniformly from $60^\circ$ to $90^\circ$;
the ignition crossing is at $C+0.5^\circ$. Success requires the
crossing event, a final angle within $3^\circ$ of $C+0.5^\circ$,
a visible flame, gripper opening of at least 80\%, and TCP height
at least 30\,mm above the knob center. Angular speed is not an
additional success gate for this task.

\paragraph{Stove Press}
The peak magnitude $C$ is sampled uniformly from $60^\circ$ to
$90^\circ$. The knob has 8\,mm of axial travel and engages after
7\,mm of depression. Success is a latched event: while engaged,
the knob must cross $d\theta\ge C+0.5^\circ$ in the easy direction.
The evaluator does not additionally require final-angle tolerance,
release, or final flame visibility. Thus, its completion logic differs
from the terminal-state conjunction used for Stove Turn.

\paragraph{Drawer Pull}
The resistance onset ranges from 45 to 90\,mm of opening, the
rising-resistance band length from 12 to 30\,mm, and breakaway
resistance from 5.5 to 9\,N. The target opening $x_{\mathrm{target}}$
corresponds to 2.35\,N of designed resistance. With onset $s$, the
accepted opening interval is
$[s,x_{\mathrm{target}}+2.5\,\mathrm{mm}]$.
Completion requires entering the resistance region, remaining within
the accepted opening range with 15 stable frames, an intact latch,
and gripper opening of at least 65\%. Crossing the opening range
without completing release and stability is not counted as success.

\paragraph{Nut Thread}
In the task's world frame, the bolt position is sampled uniformly
from $x\in[0.58,0.68]$\,m and $y\in[-0.11,0.11]$\,m, rejecting
positions within 0.14\,m of the pickup location $(0.49,0)$\,m.
Thread pitch is 2\,mm. The evaluator requires horizontal center
error at most 2.5\,mm, signed nut-base height error in
$[-2,+1]$\,mm relative to the target, and nut-axis deviation from
vertical at most $10^\circ$. The axis check is unoriented. The target
nut-base height is 32\,mm above the bolt root. Completion also requires
gripper opening of at least 65\% and clearance of the fingertip
midpoint from the nut base: either horizontal separation of at
least 70\,mm or height at least 55\,mm above the nut base.
These are operational geometric and completion checks; they do not
constitute a separate classifier of thread engagement.

\clearpage
\section{Force-Interface Implementation Details}
\label{app:interfaces}

This section specifies the feature construction and modules underlying
the designs in Sec.~\ref{sec:methods}. Input placement, normalization,
and model capacity vary across the designs; consequently, the
comparison should not be interpreted as changing only a single
insertion location.

\subsection{Numerical Prompt Features}
Prompt Fusion appends the current six-axis wrench to the seven robot
state values, giving 13 meaningful scalars. Temporal Prompt appends
the current wrench and the per-axis mean and standard deviation of
a 16-sample window, giving $7+6+6+6=25$ scalars. Both vectors are
padded to the model's state width of 32 before numerical tokenization.
The temporal prompt computes mean and standard deviation over all
entries of the supplied window; it does not apply the validity mask
used by the separate Dual-Path token encoder.

For the numerical state pathway, component-wise quantile normalization
uses the stored lower and upper quantiles:
\begin{equation}
\widetilde{x}_j =
2\frac{x_j-q_{0.01,j}}{q_{0.99,j}-q_{0.01,j}+10^{-6}}-1.
\label{eq:appendix_quantile}
\end{equation}
Here $q_{0.01,j}$ and $q_{0.99,j}$ denote the corresponding stored
normalization statistics for feature $j$. The numerical tokenizer
applies 256 equally spaced thresholds
$b_k=-1+2k/256$, $k=0,\ldots,255$.
It returns the index of the largest threshold not exceeding the
normalized value, with index $-1$ below the first threshold. The
tokenizer itself does not explicitly clip the input to $[-1,1]$.
This numerical discretization is distinct from the continuous-force
pathway, which clips its normalized force history and future-force
labels to $[-5,5]$.

\subsection{Continuous Force Tokens and Attention}
The current-wrench encoder uses a linear map from 6 to $6\times128$
values, followed by GELU. After reshaping into six hidden vectors, a
shared linear projection from 128 to 1024 dimensions and layer
normalization produce the six force tokens. The token sequence is
inserted before the action tokens in the action-expert suffix.
The force block can attend to the visual--language prefix and itself;
the action block can additionally attend to the force block. The
prefix cannot attend to the suffix.

Cross-Attention uses the same current-wrench token construction, but
conditions the action representation after the action expert and before
the action projection. Its attention module has eight heads, with
1024-dimensional query, key, value, and output representations.

\subsection{Dual-Path Statistics and Dropout}
For a 16-sample force window $w_i\in\mathbb{R}^6$ with validity
indicators $m_i$, set $n=\max(1,\sum_i m_i)$.
The masked statistics are
\begin{align}
\mu &= \frac{1}{n}\sum_i m_iw_i,\\
\sigma &= \sqrt{\frac{1}{n}\sum_i m_i(w_i-\mu)^2+10^{-6}},
\end{align}
with element-wise operations. The same masked averaging rule is
applied separately to the first and second eight samples to obtain
$\mu_{\mathrm{first}}$ and $\mu_{\mathrm{second}}$. Their denominators
are each bounded below by one. A presence indicator
$p=\mathbf{1}[\sum_i m_i>0]$ completes the feature vector
\begin{equation}
h=[\mu,\sigma,\mu_{\mathrm{first}},\mu_{\mathrm{second}},p]
\in\mathbb{R}^{25}.
\end{equation}
In particular, this vector contains no additional current-wrench
component. A linear map $25\rightarrow4\times128$, GELU, and
reshaping produce four hidden vectors, followed by the shared
$128\rightarrow1024$ projection and layer normalization.

During training, whole-window dropout zeros the history and its
validity mask for this second pathway with probability 0.3. The
already constructed numerical prompt is retained. Thus, dropout
does not remove force information from both pathways. At deployment,
the history is supplied without this training-time dropout.

\subsection{Auxiliary Future-Force Prediction}
The prediction head in Token Fusion + Pred first mean-pools the
contextualized force tokens and projects the result from 1024 to
256 dimensions with GELU. A separate projection maps each estimated
clean action from the model's 32-dimensional action representation
to 256 dimensions with GELU, then pools over the action chunk.
For noisy actions $x_t$, flow output $v_t$, and flow time $t$, the
clean-action estimate is $x_t-tv_t$; this $t$ is the flow time,
not the environment-step index.

The two pooled vectors are concatenated and passed through a
512-unit hidden layer with GELU and a 60-dimensional output layer.
Reshaping yields ten six-axis future-force predictions. The loss in
Sec.~\ref{sec:methods} excludes padded future targets and averages
over valid steps and six components, with weight $3\times10^{-4}$.
The head supplies training supervision and is not evaluated during
deployment action sampling. Pooling summarizes the chunk without
preserving its action order.

\section{Aggregation and Interpretation of Results}
\label{app:interpretation}

\subsection{Episode-Level Success and Category Means}
For method $m$, task $t$, and evaluation episode $i$, let
$y_{mti}\in\{0,1\}$ denote the complete task outcome. With $N_t$
episodes, task success in percent is
\begin{equation}
S_{mt}=\frac{100}{N_t}\sum_{i=1}^{N_t}y_{mti},\qquad
\overline{S}_m=\frac{1}{T}\sum_{t=1}^{T}S_{mt}.
\end{equation}
Simulation uses $T=8$ and $N_t=200$. Because the task sample sizes
are equal, the overall mean is also the percentage of successes
among the 1,600 episodes for one method. The ten methods in
Table~\ref{tab:preliminary} therefore account for 16,000 evaluations.

The physical-property category contains Weight Sort and CoM Sort;
the resistance category contains Faucet Turn, Faucet Direction, Stove
Turn, Stove Press, and Drawer Pull; the assembly category contains
Nut Thread. Each category mean weights its tasks equally. The overall
mean consequently weights the three category means by $2/8$, $5/8$,
and $1/8$, rather than averaging the three categories equally.

\begin{table}[t]
\centering
\caption{Simulation category means (\%) and total successes, derived
from Table~\ref{tab:preliminary}. Each total is out of 1,600 episodes.
Property, resistance, and assembly contain two, five, and one task,
respectively.}
\label{tab:appendix_categories}
\fontsize{8}{9.5}\selectfont
\setlength{\tabcolsep}{3pt}
\begin{tabular}{@{}lrrrr@{}}
\toprule
Method & Property & Resistance & Assembly & Total\\
\midrule
No Force & 49.75 & 11.00 & 31.50 & 372\\
Token Fusion & 50.25 & 12.40 & 26.00 & 377\\
Prompt Fusion & 72.25 & 20.80 & 34.00 & 565\\
Cross-Attention & 52.75 & 10.80 & 17.50 & 354\\
Token Fusion + Pred & 63.25 & 11.40 & 26.50 & 420\\
Temporal Prompt & 79.75 & 15.90 & 22.00 & 522\\
Dual-Path Fusion & 85.75 & 18.60 & 16.50 & 562\\
$\pi_0$ & 50.25 & 10.00 & 36.00 & 373\\
ForceVLA & 40.50 & 13.70 & 26.00 & 351\\
ForceFlow & 27.25 & 0.10 & 0.00 & 110\\
\bottomrule
\end{tabular}
\end{table}

An additional successful simulation episode changes a task rate by
0.5 percentage points and the overall rate by 0.0625 percentage points.
Prompt Fusion and Dual-Path Fusion achieve 565 and 562 successes,
respectively: their difference is three episodes, or 0.1875 percentage
points before rounding. ``Highest observed'' therefore denotes the
ordering of the recorded point estimates, not an established
statistically significant advantage. Category and task differences
provide additional information beyond this small overall gap.

\subsection{Real-World Study}
The hardware study uses a Franka Research 3 arm, a KWR75B six-axis
force/torque sensor, a Robotiq gripper, and external and wrist cameras.
Each system uses 200 demonstrations per task and is trained for
10,000 steps with batch size 64. Methods are evaluated from the same
20 initial conditions for each task. Across eight systems and five
tasks this gives $8\times5\times20=800$ trials, or 100 trials per
system.

The five tasks probe different manipulation demands. Weight Sort
distinguishes visually similar full and empty cans. Plate Retrieve
extracts a plate from foam; Stamp Press targets a marked region;
Plug Insert requires resistance-guided seating; and Block Insert
requires geometrically constrained insertion. In the study's category
assignment, Weight Sort is physical-property inference, Plate Retrieve,
Stamp Press, and Plug Insert are resistance-dependent interaction,
and Block Insert is assembly.

The real-world mean averages the five task rates equally. One
successful trial changes a task rate by 5 percentage points and the
overall mean by 1 percentage point. Prompt Fusion and Dual-Path
Fusion each achieve 63 successes out of 100, compared with 40 for
No Force. Their equal totals conceal different task profiles:
Prompt Fusion improves all five tasks relative to No Force, whereas
Dual-Path improves Weight Sort from 6 to 19 successes but reduces
Block Insert from 10 to 4 successes out of 20. These counts are
equivalent to the percentages in Table~\ref{tab:hardware_results},
not additional experiments.

\subsection{Diagnostic and Statistical Scope}
Intermediate-state diagnostics do not replace complete-task success.
For example, Prompt Fusion and Dual-Path Fusion each reach the
Drawer Pull opening range with an intact latch in 81 episodes, but
their complete successes are 46 and 9, respectively. The discrepancy
reflects the remaining completion conditions. Similarly, sorting
destination choices and assembly target geometry should be reported
separately from the final release and withdrawal checks.

Shared environment seeds and initial conditions support comparable
evaluations, but the aggregate success tables alone do not encode
which paired episodes differ. They also do not quantify variability
across independently trained policies. We therefore interpret the
reported values as observed performance under the stated protocol,
without deriving paired significance tests or training-seed variance
from these aggregate counts.

The controlled comparison uses one VLA backbone, eight simulation
tasks, and one simulated robot embodiment. The assembly category has
one task, so its category-level result should not be generalized to
all assembly problems. The real-world study has 20 trials per
task--method pair and supports qualitative comparisons across the
tested tasks; it does not establish a general simulation-to-reality
ranking correlation. The external reference systems also differ in
architecture, training budget, and, for ForceFlow, execution horizon.
Their scores describe the evaluated systems under those settings.

\section{Example Six-Axis Wrench Trajectories}
\label{app:wrench}

\begin{figure*}[!t]
\centering
\includegraphics[width=\textwidth]{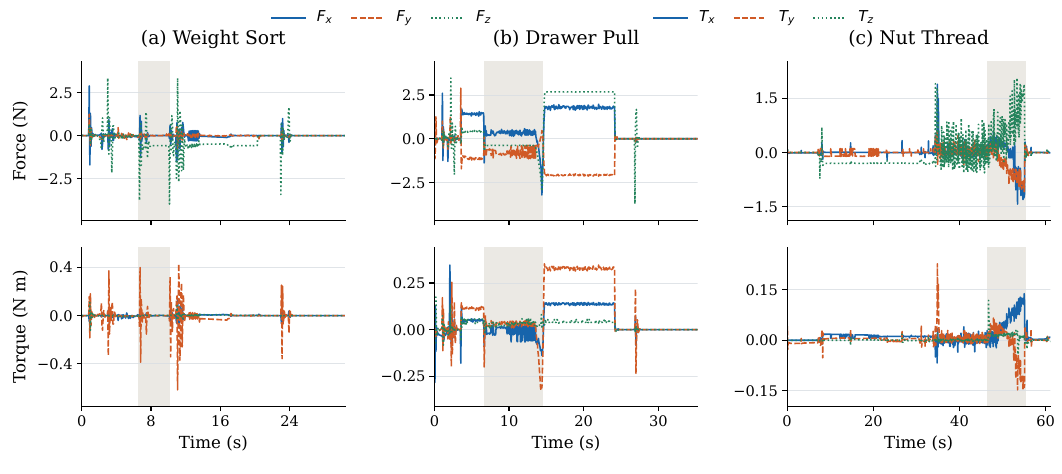}
\caption{Six-axis wrench trajectories from three successful simulation
demonstrations, each selected at seed 0 (episode 1). Top: $F_x,F_y,F_z$
in newtons. Bottom: $T_x,T_y,T_z$ in newton-metres. Gray intervals mark
the logged lift-and-weigh phase in (a), resistance-rise phase in (b),
and thread-engagement phase in (c). All 457, 530, and 917 control-rate
samples are shown, respectively. These are illustrative expert
trajectories, not comparisons between learned policies.}
\label{fig:wrench_sim_examples}
\end{figure*}

\begin{figure*}[!t]
\centering
\includegraphics[width=\textwidth]{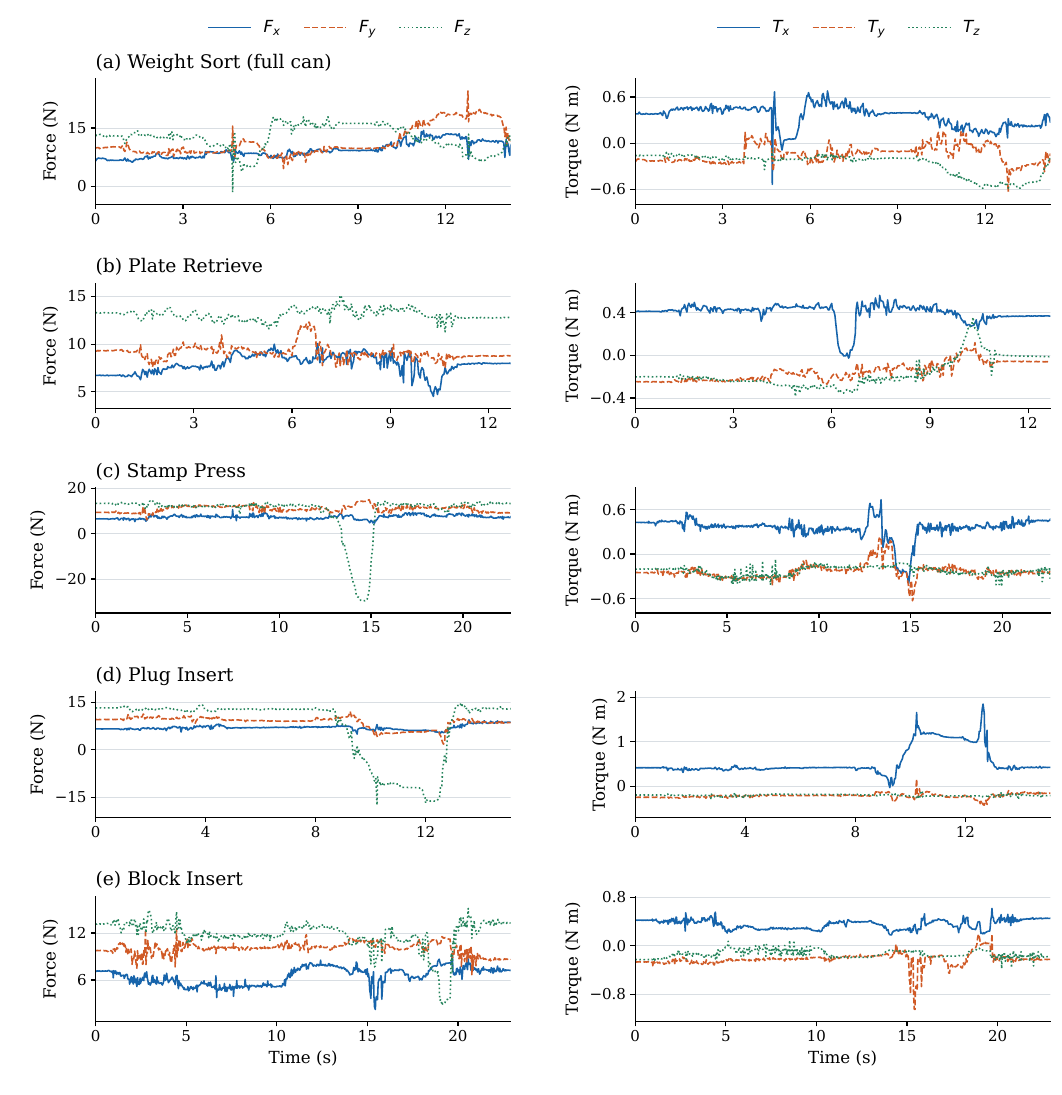}
\caption{Six-axis wrench trajectories covering all five real-world
tasks. Each row shows one recorded demonstration: (a) Weight Sort
(full can), (b) Plate Retrieve, (c) Stamp Press, (d) Plug Insert, and
(e) Block Insert. Left: $F_x,F_y,F_z$ in newtons. Right: $T_x,T_y,T_z$
in newton-metres. All 424, 376, 675, 442, and 684 samples are retained,
respectively, including offsets and transient variations. Time is
relative to the first recorded sample; vertical scales vary by task
and quantity. Colors and line styles follow
Fig.~\ref{fig:wrench_sim_examples}. The source records are episodes
0, 0, 1, 0, and 0 of the corresponding archived task collections,
and are illustrative demonstrations rather than policy-evaluation
rollouts.}
\label{fig:wrench_real_examples}
\end{figure*}

Figures~\ref{fig:wrench_sim_examples} and~\ref{fig:wrench_real_examples}
show the temporal structure of the six-axis observations in example
task recordings. Force and torque components are plotted on separate
axes. We plot every recorded sample without
smoothing, normalization, clipping, or subtraction of the initial
reading. Only the time origin is shifted to the first recorded sample;
axis ranges are chosen separately for each task and quantity.

The simulation examples are the seed-0, episode-1 successful scripted
demonstrations for Weight Sort, Drawer Pull, and Nut Thread. They use
the raw 15\,Hz control-rate wrist--hand joint reaction wrench described
in Appendix~\ref{app:simulation}, rather than the separate 60\,Hz
physics stream. Shaded intervals come from recorded expert-phase
labels, not from detecting events in the plotted signal.

The hardware examples cover all five tasks: Weight Sort (a full-can
demonstration), Plate Retrieve, Stamp Press, Plug Insert, and Block
Insert. Their six-axis KWR75B readings are
expressed in the sensor frame, with no coordinate transformation or
bias compensation recorded in the acquisition metadata. The horizontal
axes use the actual timestamps; the selected records have average
sampling rates of approximately 29.2--29.8\,Hz. These examples illustrate
recorded observations rather than additional trials in the 800-trial
policy evaluation. Joint-reaction and sensor-frame readings have
different reference conventions and are not isolated contact-force
estimates; their absolute magnitudes should not be used as a calibrated
simulation-to-reality comparison.

\end{document}